\documentclass[10pt,twocolumn,letterpaper]{article}

\usepackage[accsupp]{axessibility}  % Improves PDF readability for those with disabilities.

\usepackage{cvpr}      % To produce the REVIEW version
\usepackage{graphicx}
\usepackage{amsmath}
\usepackage{amssymb}
\usepackage{booktabs}
\usepackage{stfloats}
\usepackage{multirow}
\usepackage{url}
\usepackage[dvipsnames]{xcolor}
\usepackage{xcolor}
\usepackage[accsupp]{axessibility}  % Improves PDF readability for those with disabilities.

\usepackage[pagebackref,breaklinks,colorlinks]{hyperref}

\usepackage[capitalize]{cleveref}
\crefname{section}{Sec.}{Secs.}
\Crefname{section}{Section}{Sections}
\Crefname{table}{Table}{Tables}
\crefname{table}{Tab.}{Tabs.}

\def\cvprPaperID{4814} % *** Enter the CVPR Paper ID here
\def\confName{CVPR}
\def\confYear{2023}

\definecolor{DarkGreen}{rgb}{0.0, 0.5, 0.0}

\begin{document}

% \title{ULIP: Learning Universal Representation of Image, Text and Point Cloud}
%\title{Multi-Modality Pre-Training Can Help 3-D Recognition}
\title{ULIP: Learning a Unified Representation of Language, Images, and Point Clouds for 3D Understanding}
\author{Le Xue$^{1}$\thanks{\ Contact: lxue@salesforce.com},
% {\tt\small lxue@salesforce.com}
Mingfei Gao$^{1}$,
% {\tt\small mingfei.gao@salesforce.com}
Chen Xing$^{1}$,
% {\tt\small cxing@salesforce.com}
Roberto Martín-Martín$^{1, 3}$,
% {\tt\small rmartinmartin@salesforce.com}
Jiajun Wu$^{2}$,
% {\tt\small jiajunwu@cs.stanford.edu}
Caiming Xiong$^{1}$,
% {\tt\small cxiong@salesforce.com}
\\
Ran Xu$^{1}$,
% {\tt\small ran.xu@salesforce.com}
Juan Carlos Niebles$^{1, 2}$, and
% {\tt\small jniebles@salesforce.com}
Silvio Savarese$^{1, 2}$
\\
\\
% {\tt\small ssavarese@salesforce.com}\\
$^{1}$ Salesforce AI, Palo Alto, USA \\
$^{2}$ Stanford University, Stanford, USA \quad ${^3}$ UT Austin, Texas, USA \\
% \\
% {\tt\small jiajunwu@cs.stanford.edu} \\
% {\tt\small{\{lxue, mingfei.gao, cxing, rmartinmartin, cxiong, ran.xu, jniebles, ssavarese\}@salesforce.com}} \\
{\tt\small Project Website: \href{https://tycho-xue.github.io/ULIP/}{https://tycho-xue.github.io/ULIP/}}}

\maketitle

%%%%%%%%% ABSTRACT
\begin{abstract}
%\jw{Re title: maybe say "Image, Text" instead of "Language, Image" for consistency across the entire paper.}
% A unified text and image representation space, such as that of CLIP, has significantly enhanced various image recognition tasks. However, limited progress has been made in creating a multi-modal representation space that involves 3D modality. 
% State-of-the-art computer vision systems are
% trained to predict a fixed set of pre-determined
% object categories. This restricted form of supervision limits their generality and usability since
% additional labeled data is needed to specify any
% other visual concept. Learning directly from raw
% text about images is a promising alternative which
% leverages a much broader source of supervision.
%This restricted form of supervision limits their generality and usability since additional labeled data is needed to specify any other visual concept.
The recognition capabilities of current state-of-the-art 3D models are limited by datasets with a small number of annotated data and a pre-defined set of categories.
In its 2D counterpart, recent advances have shown that similar problems can be significantly alleviated by employing knowledge from other modalities, such as language.
Inspired by this, leveraging multimodal information for 3D modality could be promising to improve 3D understanding under the restricted data regime, but this line of research is not well studied.
Therefore, we introduce ULIP to learn a unified representation of images, texts, and 3D point clouds by pre-training with object triplets from the three modalities.
To overcome the shortage of training triplets, ULIP leverages a pre-trained vision-language model that has already learned a common visual and textual space by training with massive image-text pairs. Then, ULIP learns a 3D representation space aligned with the common image-text space, using a small number of automatically synthesized triplets.
ULIP is agnostic to 3D backbone networks and can easily be integrated into any 3D architecture.
Experiments show that ULIP effectively improves the performance of multiple recent 3D backbones by simply pre-training them on ShapeNet55 using our framework, achieving state-of-the-art performance in both standard 3D classification and zero-shot 3D classification on ModelNet40 and ScanObjectNN. ULIP also improves the performance of PointMLP by around 3\% in 3D classification on ScanObjectNN, and outperforms PointCLIP by 28.8\% on top-1 accuracy for zero-shot 3D classification on ModelNet40. Our code and pre-trained models are released at \href{https://github.com/salesforce/ULIP}{https://github.com/salesforce/ULIP}.
\end{abstract}

% some comments from Roberto: "based on the current results we have, ideal case we want to claim we try to align three modalities into one feature space, but might be challenged for not enough experiments or freeze the CLIP weights in pre-train, on the other hand, the most conservative claim is that we use CLIP to help 3d, it will sound less exciting, but it will sound more exciting if we try to clain we make a  framework where you can align third modality into CLIP's existing feature space and demonstrate the benefits of aligning a third modality into CLIP image-text space. Another recommendation is to think maybe we can meet in the middle for the two solutions, since the CLIP's data are much better compared to 3d, since the CLIP image-text space is pretty well learnt and powerful do we really need to train these 3 modalities together again, our goal is to align 3 modalities together, but based on what we got for the 3d dataset, we are actually taking advantage of CLIP and adopt a 'smart' way to achieve aligning 3 modalities together by align 3d modality into the image-text space instead."
%Also, we might need to think how to present the hard sets in modelnet40 for zero-shot classification, since it's manually determined, it won't affect the conclusion though.

%%%%%%%%% BODY TEXT
\section{Introduction}
\label{sec:intro}

%1.1 talk about the importance of 3d vision, usefulness, benefits, but also emphasize the current constraints and difficulties. (Maybe mention pre-training is a way to help with the constraints to show it's one of our motivations. Maybe also mention other modalities interaction with 3d in the pre-training phase has not been well studied as well. maybe mention existing multimodal 3d work, kind of still rely on conversion 3d to 2d.)
%1.2 to address the limitations of 3d vision from the data side, talk about self supervised learning success, and some existing success in the 2d/language space, also some pre existing work in 3d SSL, but mention the limitations of current 3d SSL works. (In contrast multimodal pre-training can help with the limitations, it's also one of our motivations.)
%1.3 talk about benefits and breakthroughs of multimodal pre-training, emphasize that it enables many multimodal applications, make high accuracy zero-shot classification possible, and yield good pre-trained weights for downstreaming tasks.
%1.4 talk about some current multimodal 3d works, emphasize the drawbacks like they still need to convert 3d to 2d, then use 2d methods, thus bring up study on multimodal pre-training for 3d is meaningful and potentially can address the existing problems.
 
\begin{figure}[hbt]
    \centering
    \includegraphics[width=1.0\linewidth]{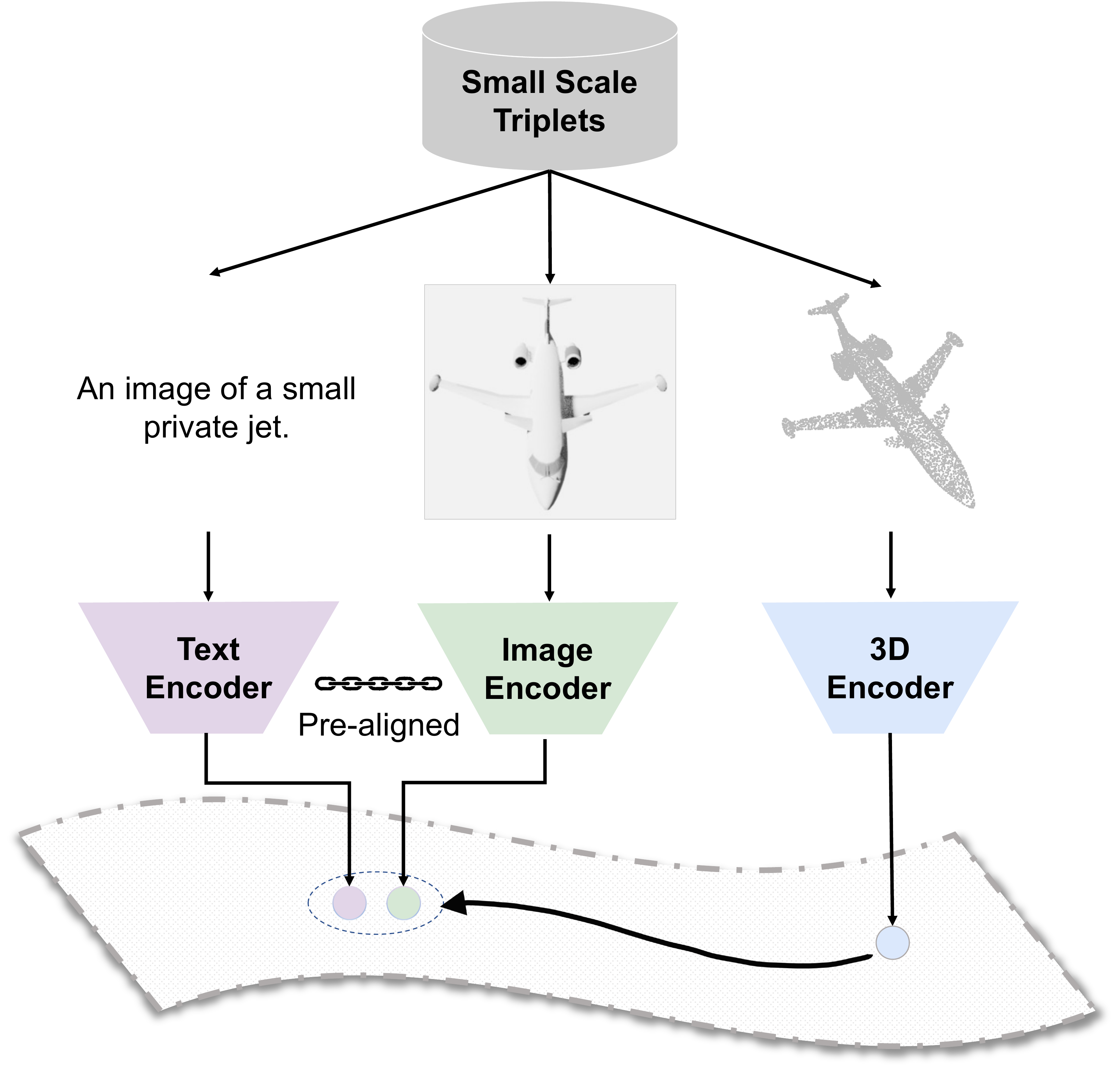}
    \caption{Illustration of ULIP. ULIP improves 3D understanding by aligning features from images, texts, and point clouds in the same space. To reduce the demand of 3D data, ULIP leverages image and text encoders that are pre-trained with large-scale image-text pairs, and aligns 3D representation to the pre-aligned image-text feature space using a small scale of training triplets. }
    \label{fig:alignment}
     %\vspace{-5mm}
\end{figure}

3D visual understanding research~\cite{landrieu2018large, hu2020randla, liu2019densepoint, li2021lidar, graham20183d, choy20163d} is drawing significant attention in recent years due to the increasing demand of real-world applications such as augmented/virtual reality~\cite{vu2022softgroup, liu2021group, misra2021end, armeni20163d}, autonomous driving~\cite{yin2021center, li2022deepfusion} and robotics~\cite{wojek2011monocular, cadena2016multi}.
However, compared to their 2D counterpart, 3D visual recognition research is still limited by datasets with a small number of samples and a small set of pre-determined categories~\cite{Uy_2019_ICCV, wu20153d}.
For example, ShapeNet55~\cite{chang2015shapenet}, one of the largest publicly available 3D datasets, only contains around 52.5k samples of 3D objects with 55 category labels. 
That is in contrast to the 2D domain, where ImageNet~\cite{deng2009imagenet} contains millions of images that cover thousands of categories.
This scale limit of 3D data, caused by the high cost of 3D data collection and annotation~\cite{yu2022point, chang2015shapenet, goyal2021revisiting, wu20153d}, has been hindering the generalization of 3D recognition models and their real-world applications.

To tackle the shortage of annotated data,  existing work in other domains shows that employing knowledge from different modalities can significantly help the concept understanding in the original modality\cite{radford2021learning,xing2019adaptive}. 
Among such work, CLIP~\cite{radford2021learning} pioneered alignment between visual and textual features by pre-training on large-scale image-text pairs. It improves state-of-the-art visual concept recognition and enables zero-shot classification of unseen objects.
However, multimodal learning that involves 3D modality, and whether it can help 3D recognition tasks are still not well studied.

In this paper, we propose Learning a \textbf{U}nified Representation of \textbf{L}anguage, \textbf{I}mages, and \textbf{P}oint Clouds (ULIP). 
An illustration of our framework is shown in Figure \ref{fig:alignment}. 
Obtaining a unified representation space of all three modalities requires large-scale triplets of image, text, and point cloud as training data.
However, such triplets remain hard to collect compared to the large-scale image-text pairs available.
To circumvent the lack of triplet data, we take advantage of a vision-language model pretrained on massive image-text pairs, and align the feature space of a 3D point cloud encoder to the pre-aligned vision/language feature space. 
When training the 3D encoder for space alignments, we use a small number of automatically synthesized triplets from ShapeNet55~\cite{chang2015shapenet} without requiring manual annotations.
Making use of a pretrained vision-language model lets us leverage the abundant semantics captured in the image-text feature space for 3D understanding.
%Specifically, for an arbitrary 3D backbone encoder, we pre-train it on object triplets (image, text, and point cloud). 
Our framework uses CLIP as the vision and language model because of its excellent generalization performance. During pre-training, we keep the CLIP model frozen and train the 3D encoder by aligning the 3D feature of an object with its corresponding textual and visual features from CLIP using contrastive learning.
The pre-trained 3D backbone model can be further fine-tuned for different downstream tasks.

% However, multi-modal learning that involves 3D modality is still not well studied. We fill this research gap and propose a efficient pre-training framework, called ULIP, that unifies features of image, text and point cloud at one-shot. We take advantage of the powerful CLIP model and align the feature of 3D backbones to the CLIP space via contrastive learning. Our framework has following major advantages: (1) 3D backbone can be greatly improved with the knowledge distilled from CLIP; (2) our framework is agnostic to 3D network. Therefore, we can easily plug in any 3D backbone and improve it and (3) compared to two modalities, aligning three modalities in the same space can potentially enable more cross-domain downstream tasks including zero-shot 3D classification and text-3D/image-3D retrieval.

ULIP has three major advantages. First, ULIP can substantially improve the recognition ability of 3D backbone models. Second, ULIP is agnostic to the architecture of 3D models; therefore, we can easily plug in any 3D backbones and improve them with ULIP. Third, aligning three modalities in the same feature space can potentially enable more cross-domain downstream tasks, including zero-shot 3D classification and image-to-3D retrieval.

We quantitatively evaluate ULIP on two fundamental 3D tasks: standard 3D classification and zero-shot 3D classification. We experiment with recent 3D networks including PointNet++ ~\cite{qi2017pointnet++}, PointMLP~\cite{ma2022rethinking} and PointBERT~\cite{yu2022point}. 
Experimental results show that ULIP achieves state-of-the-art (SOTA) performance for both standard 3D classification and zero-shot 3D classification on ModelNet40 and ScanObjectNN. Specifically, ULIP surpasses PointMLP by around 3\% in standard 3D classification on ScanObjectNN~\cite{Uy_2019_ICCV}. ULIP also outperforms PointCLIP~\cite{zhang2022pointclip} (the previous SOTA) by around 28.8\% top-1 accuracy in zero-shot 3D classification on ModelNet40. Moreover, we showcase the potential of applying ULIP on the image to point cloud retrieval task. %, where we use real images from Caltech101 as queries and retrieve the top 5 point clouds from all candidates in ModelNet40. 
Qualitative evaluation demonstrate our promising potential for cross-modal applications.

% Our contributions are summarized as follows.
% \begin{itemize}
%     \item We propose ULIP, a multi-modality pre-training framework that improves standard 3d classification models with the help of 2D vision and language modalities.
%     \item ULIP is a general architecture agnostic to architecture of 3D backbones. It can be applied to improve different 3D networks without modification.
%     \item ULIP sets the new state-of-the-art performance in standard 3D classification and zero-shot 3D classification in Modelnet40 and ScanobjectNN datasets.
% \end{itemize}

% Our contributions are summarized as follows.
% \begin{itemize}
%     \item We propose ULIP, a simple yet effective pre-training framework that improves 3D backbones with the help of 2D vision and language modalities.
%     \item ULIP is a general architecture that is agnostic to the architecture of 3D backbones. So, it can be used to improve different 3D networks without modification.
%     \item ULIP sets the new state-of-the-art performance in standard 3D classification and zero-shot 3D classification in Modelnet40 and ScanobjectNN datasets.
% \end{itemize}

%------------------------------------------------------------------------
\section{Related Work}
\label{sec:related work}
\noindent\textbf{Multi-modal Representation Learning.}
Most existing multimodal approaches are about image and text modalities. 
Among these methods, one line of research focuses on learning interaction between image regions and caption words \cite{tan2019lxmert,chen2020uniter,li2020oscar,lu2019vilbert,li2019visualbert,li2021align} using transformer-based architectures. These methods show great predictive capability while being costly to train. 
The other line of research, such as CLIP \cite{radford2021learning}, uses image and text encoders to output a single image/text representation for each image-text pair, and then aligns the representations from both modalities. This simple architecture makes training with massive noisy web data efficient, facilitating its zero-shot generalization capability.

The success of CLIP has promoted many image-text related research directions, including text-based image manipulation \cite{patashnik2021styleclip}, open vocabulary object detection \cite{gu2021open,gao2021towards} and language grounding \cite{li2022grounded}. Some recent works explore how multi-modal information can help 3D understanding and show promising results \cite{yan2022let, chen2021multimodal}. The most related method to our work is PointCLIP \cite{zhang2022pointclip}. It first converts the 3D point cloud into a set of depth maps and then leverages CLIP directly for zero-shot 3D classification. Unlike PointCLIP, which targets reshaping the task of point cloud and text matching to image and text alignment, our method learns a unified representation among images, texts, and point clouds that substantially improves 3D understanding.

\noindent\textbf{3D Point Cloud Understanding.}
There are mainly two streams of research lines for point cloud modeling. One is projecting a point cloud into 3D voxels \cite{maturana2015voxnet,shi2020pv} and then using 2D/3D convolutions for feature extraction. PointNet \cite{qi2017pointnet} explores ingesting 3D point clouds directly. It extracts permutation-invariant feature from the point cloud that significantly impacts point-based 3D networks. PointNet++ \cite{qi2017pointnet++} proposes a hierarchical neural network that extracts local features with increasing contextual scales. Recently, PointMLP \cite{ma2022rethinking} proposes a pure residual MLP network and achieves competitive results without integrating sophisticated local geometrical extractors. 
Moreover, self-supervised learning for 3D point clouds has also shown promising performance in 3D understanding field. PointBERT \cite{yu2022point} adopts mask language modeling from BERT \cite{devlin2018bert} to the 3D field, where it tokenizes 3D patches using an external model, randomly masks out 3D tokens, and predicts them back during pre-training. A more recent work, PointMAE \cite{pang2022masked}, directly operates the point cloud by masking out 3D patches and predicting them back using L2 loss. Our method is orthogonal to the above 3D encoders. Their performance on 3D recognition can be potentially improved by ULIP with no/minor modification.

%------------------------------------------------------------------------
\section{Learning a Unified Representation of Language,
Images, and Point Clouds}
\label{sec:method}
% ULIP learns a universal representation  via pre-training on large-scale triplets of image, text, and point cloud. So, we first introduce the dataset and cost efficient strategy we utilized to create training triplets. Then, we present our pre-training framework.

ULIP learns a unified representation space of language, images, and 3D point clouds via pre-training on triplets from these three modalities. 
In this section, we first introduce how we create such triplets for pre-training. Then, we present our pre-training framework.

%-------------------------------------------------------------------------
\subsection{Creating Training Triplets for ULIP}
\label{sec:create triplet}
% ShapeNet55 [cite] is one of the largest public 3D point cloud datasets. It contains around 55K CAD models and each of which is associated with meta data that textually describe the semantic information of each CAD model. For each CAD model in the dataset, we create a triplet $T_i:(I_i, S_i, P_i)$ of image, text (sentence) and point cloud. Our model is pre-trained using those triplets.

We build our dataset of triplets from ShapeNet55~\cite{chang2015shapenet}, which is one of the most extensive public 3D CAD datasets. 
ShapeNet55 is the publicly-available subset of ShapeNet. 
It contains around 52.5K CAD models, each of which is associated with metadata that textually describes the semantic information of the CAD model. 
For each CAD model $i$ in the dataset, we create a triplet $T_i:(I_i, S_i, P_i)$ of image $I_i$, text description $S_i$ and point cloud $P_i$. ULIP will then use these triplets for pre-training. 

%\jw{here it becomes hard to understand. I think we should refer to Figure 2.}
\noindent\textbf{Point Cloud Generation}. We directly use the generated point cloud of each CAD model in ShapeNet55. We uniformly sample $N_p$ points from the original point cloud. During pre-training, standard data augmentation techniques of 3D point clouds are performed, including random point drop, random scaling point cloud, shift point cloud and rotate perturbation.
Then a 3D encoder takes the augmented point cloud $P_i$ as input and outputs its 3D representation $\mathbf{h}_i^{P}$ via
\begin{equation}
 \label{eq:3d-rep}
   \mathbf{h}_i^{P} = f_P(P_i),
\end{equation}
where $f_P(\cdot)$ represents the 3D backbone encoder.

\begin{figure*}[t]
    \centering
    \includegraphics[width=1.0\linewidth]{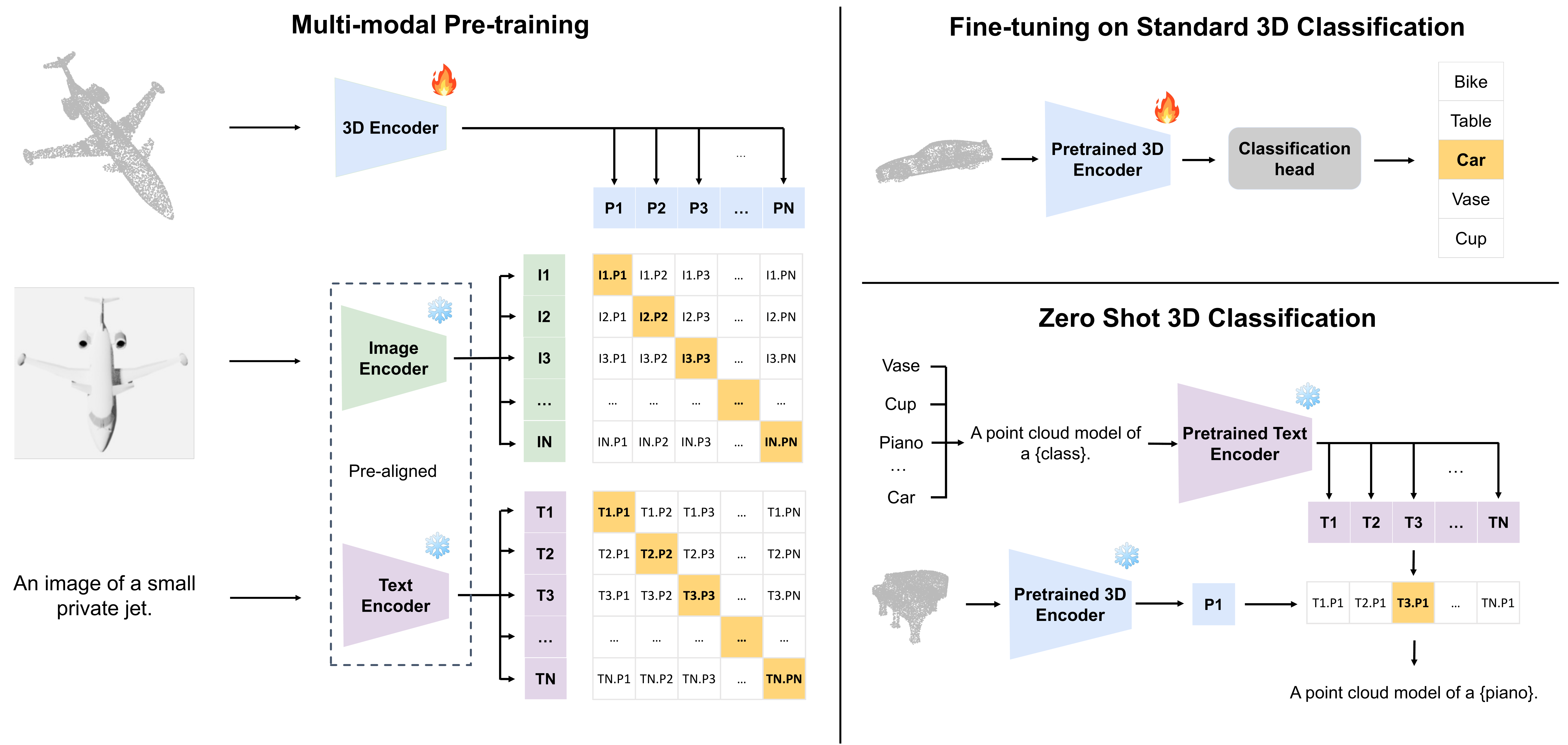}
    %\vspace{-5pt}
    \caption{Illustration of our method. The inputs of multimodal pre-training (\textbf{Left}) are a batch of objects represented as triplets (image, text, point cloud). Image and text features are extracted from a pre-trained (frozen) vision and language model such as CLIP, and 3D features are extracted from a 3D encoder. Contrastive losses are applied to align the 3D feature of an object to its image and text features during pre-training. The pre-trained 3D encoders are further fine-tuned in downstream tasks, including standard 3D classification (\textbf{Top Right}) and zero-shot 3D classification (\textbf{Bottom Right}).}
    \label{fig:method}
    %\vspace{-6mm}
\end{figure*}

\noindent\textbf{Multi-view Image Rendering}. ShapeNet55 CAD models do not come with images. 
To obtain images that semantically align well with each CAD model, we synthesize multi-view images of each CAD model by placing virtual cameras around each object and rendering the corresponding RGB images and depth maps from each viewpoint.\footnote{We utilize the following repository with their default settings in practice. \newline https://github.com/panmari/stanford-shapenet-renderer} Specifically, we render an RGB image with a depth map for every 12 degrees. Therefore, we get 30 RGB images and 30 depth maps for each object, 60 image candidates in total. During each iteration of pre-training, we randomly select one image or depth map from each CAD model's 60 renderred candidates as $I_i$ and take $I_i$ as input of the image encoder $f_I(\cdot)$ to extract the image feature $\mathbf{h}_i^I$,
\begin{equation}
 \label{eq:image-rep}
   \mathbf{h}_i^{I} = f_I(I_i).
\end{equation}

% \noindent\textbf{Multi-view Image Rendering}. ShapeNet55 CAD models do not come with images. Obtaining real images that semantically align well with each CAD model is challenging. For simplicity, we synthesize multi-view images of each CAD model by placing virtual cameras around each object and rendering the corresponding RGB images and depth maps from each view point.\footnote{https://github.com/panmari/stanford-ShapeNet55-renderer/} Specifically, we render a RGB image with a depth map for every 12 degrees, so in total, we get 30 RGB images and 30 depth maps for each object, overall 60 image modality candidates. During each iteration of pre-training, we randomly select one image or depth map from each object and use the feature from the image encoder as the representation of image modality.

\noindent\textbf{Text Generation}. We leverage the metadata that comes with each CAD model as the corresponding text description. The metadata includes a synset of taxonomy as a textual description of each CAD model. For each word in the metadata, we adopt simple prompts to construct meaningful sentences that will be utilized during pre-training. We follow prior works~\cite{gu2021open,gao2021towards} that use 63 prompts such as "a picture of [WORD]" in image-text pre-training tasks and additionally add a dedicated prompt ``a point cloud model of [WORD]" to accommodate the 3D modality. 
In each training iteration, we randomly choose a word from the metadata and apply the 64 templates on the word to build a set of text descriptions, $\mathbf{S}_i$.
Then we input $\mathbf{S}_i$ into our text encoder $f_S(\cdot)$ and get a set of representations, respectively. Finally, we conduct average pooling over the set of outputs as the text-domain representation $\mathbf{h}_i^{S}$ of object $i$,
\begin{equation}
 \label{eq:text-rep}
   \mathbf{h}_i^{S} = Avg(f_S(\mathbf{S}_i)).
\end{equation}

%-------------------------------------------------------------------------
% \subsection{Learning Cross-modal Feature Representation}
\subsection{Aligning Representations of Three Modalities}
\label{sec:align representation}
With the created triplets of image, text, and point cloud, ULIP conducts pre-training to align representations of all three modalities into the same feature space.
Specifically, we take advantage of pre-trained vision-language models, i.e., CLIP, and train a 3D encoder by aligning the 3D feature with the features of image and text encoders ($f_I(\cdot)$ and  $f_S(\cdot)$) of CLIP.
By doing so, we hope that the abundant semantics already captured and aligned by CLIP's encoders can be employed for better 3D understanding.
The resulting unified feature space enables numerous cross-modal applications among these three modalities and potentially improves the 3D recognition performance of the underlying 3D backbone encoder $f_P(\cdot)$.

% Different from CLIP, which aligns image and text into one feature space, our goal is to add one more modality for 3D point cloud, and align the 3D feature into the image-text feature space from CLIP. In this way, we unify image, text, 3D modalities into one feature space, which will not only enable numerous multi-modal applications among these three modalities but can also potentially improve our 3D backbone while distilling knowledge from CLIP.

\noindent\textbf{Cross-modal Contrastive Learning.}
As shown in Figure \ref{fig:method}, for an object $i$, features $\mathbf{h}^I_i$, $\mathbf{h}^S_i$ and $\mathbf{h}^P_i$ are extracted from image, text, and 3D point cloud encoders. Then contrastive loss among each pair of modalities is computed as follows, 
\begin{equation}
\begin{aligned}
 L_{(M1, M2)} =
   \sum_{(i,j)}-\frac{1}{2}log\frac{\exp\left(\frac{\mathbf{h}^{M_1}_i \mathbf{h}^{M_2}_j}{\tau}\right)}{\sum_k \exp\left(\frac{\mathbf{h}^{M_1}_i \mathbf{h}^{M_2}_k}{\tau}\right)} \\-\frac{1}{2}
    log\frac{\exp\left(\frac{\mathbf{h}^{M_1}_i \mathbf{h}^{M_2}_j}{\tau}\right)}{\sum_k \exp\left(\frac{\mathbf{h}^{M_1}_k \mathbf{h}^{M_2}_j}{\tau}\right)},
\end{aligned}
\end{equation}
% \begin{equation}
% \begin{aligned}
% L_{(M1, M2)} = \sum_{(i,j)}& -\frac{1}{2} \log \frac{\exp\left(\frac{\mathbf{h}^{M_1}_i \mathbf{h}^{M_2}_j}{\tau}\right)}{\sum_k \exp\left(\frac{\mathbf{h}^{M_1}_i \mathbf{h}^{M_2}_k}{\tau}\right)} \\ -\frac{1}{2} \log \frac{\exp\left(\frac{\mathbf{h}^{M_1}_i \mathbf{h}^{M_2}_j}{\tau}\right)}{\sum_k \exp\left(\frac{\mathbf{h}^{M_1}_k \mathbf{h}^{M_2}_j}{\tau}\right)},
% \end{aligned}
% \end{equation}
where $M_1$ and $M_2$ represent two modalities and $(i, j)$ indicates a positive pair in each training batch. We use a learnable temperature parameter $\tau$ as well, similar to CLIP \cite{radford2021learning}.

Finally, we minimize $ L_{(M_1, M_2)}$ for all modality pairs with different coefficients, 
\begin{equation}
 \label{eq:final loss}
   L_{final} = \alpha L_{(I,S)}+ \beta L_{(I,P)} + \theta L_{(P,S)}.
\end{equation}
By default, $\alpha$ is set to be constant 0, $\beta$ and $\theta$ are set to be 1 equally; because during pre-training, we find that if we update CLIP's image and text encoders, catastrophic forgetting will emerge due to our limited data size. This will lead to a significant performance drop when applying ULIP to downstream tasks. Therefore we freeze the weights of $f_S(\cdot)$ and $f_I(\cdot)$ during the entire pre-training and only update $f_P(\cdot)$ with $L_{final}$.

%-------------------------------------------------------------------------
\section{Experiments}
\label{sec:experiments}
To demonstrate the benefits of pre-training 3D backbone networks using ULIP, we conduct experiments on two 3D tasks: a standard 3D classification task that involves a single modality and a zero-shot 3D classification task that involves multimodal inputs.
In this section, we first present experimental settings, including our experimenting 3D backbones, downstream datasets, and implementation details.
Then we present the quantitative results of standard 3D classification and zero-shot 3D classification, respectively. Lastly, we include analyses of our model and show results on cross-modal retrieval.

\subsection{3D Backbone Networks}
% introduce pointnet++, pointMLP, pointbert here
We experiment with the following 3D backbone networks under our framework.

\noindent\textbf{PointNet++} \cite{qi2017pointnet++} is an advanced version of PointNet \cite{qi2017pointnet}. It uses a hierarchical structure to better capture the local geometry of the point cloud, and becomes the cornerstone of many point cloud applications.
%[TODO: pointnet++ msg ssg]

\noindent\textbf{PointBERT}~\cite{yu2022point} utilizes a transformer architecture for point cloud feature extraction. It improves its recognition ability by conducting self-supervised pre-training on ShapeNet55.

\noindent\textbf{PointMLP}~\cite{ma2022rethinking} is the SOTA method on standard 3D classification task. It uses a residual MLP network with a lightweight geometric affine module to better capture local geometric features.

\begin{table}[htb]
    \small
    \centering
    \begin{tabular}{lcc}
    \toprule
         Model& Overall Acc & Class-mean Acc \\
         \midrule
         PointNet \cite{qi2017pointnet} &  68.2 & 63.4 \\
         %\hline
         PointNet++ \cite{qi2017pointnet++} &  77.9 & 75.4 \\
         %\hline
         DGCNN \cite{wu2018dgcnn} &  78.1 & 73.6 \\
         %\hline
         MVTN \cite{hamdi2021mvtn} &  82.8 &  --\\
         %\hline
         PointBERT \cite{yu2022point} &  83.1 &  --\\
         %\hline
         RepSurf-U \cite{ran2022surface} & 84.6 &  --\\
         %\hline
         PointMAE \cite{pang2022masked} & 85.2 &  --\\
         %\hline
         RepSurf-U (2x) \cite{ran2022surface} &  86.0 &  --\\
         \midrule
         %hline
         %\hline
         PointBERT \cite{yu2022point} &  83.1 &  --\\
         %\hline
         PointBERT + ULIP &  \textbf{86.4} \textcolor{DarkGreen}{\small ($\uparrow 3.3$)} & --\\
         %\hline
         \midrule
        %  pointMLP & reproduced & 1k & 85.22 & 84.13 \\
        %  \hline
         PointMLP \cite{ma2022rethinking} &  85.7 & 84.4 \\
         %\hline
         PointMLP+ ULIP &  \textbf{88.8} \textcolor{DarkGreen}{\small ($\uparrow 3.1$)} & \textbf{87.8} \textcolor{DarkGreen}{\small ($\uparrow 3.4$)} \\
         %\hline
         \midrule
         PointMLP \dag &  86.5 & 85.1 \\
         %\hline
         PointMLP \dag+ ULIP &  \textbf{89.4} \textcolor{DarkGreen}{\small ($\uparrow 2.9$)} & \textbf{88.5} \textcolor{DarkGreen}{\small ($\uparrow 3.4$)} \\
         %\hline
         \bottomrule
    \end{tabular}
    \caption{3D classification results on ScanObjectNN. ULIP significantly improves our baselines. Our best result outperforms SOTA largely by around 3\% on Overall Acc. \dag  indicates a model uses 2K sampled points and all others use 1K sampled points.}
    \label{tab:fintune-scan}
    \vspace{-4mm}
\end{table}
%2k points

\subsection{Downstream Datasets}
%introduce dataset details here
We use the following two datasets for both standard and zero-shot 3D classification.

\noindent\textbf{ModelNet40} is a synthetic dataset of 3D CAD models. It contains 9,843 training samples and 2,468 testing samples, covering 40 categories.\footnote{For each CAD model, we utilized preprocessed point cloud from \cite{ma2022rethinking}.}

\noindent\textbf{ScanObjectNN} is a dataset of scanned 3D objects from the real world.
It contains 2,902 objects that are categorized into 15 categories. It has three variants: \emph{OBJ\_ONLY} includes ground truth segmented objects extracted from the scene meshes datasets; \emph{OBJ\_BJ} has objects attached with background noises and \emph{Hardest} introduces perturbations such as translation, rotation, and scaling to the dataset\cite{Uy_2019_ICCV}.\footnote{We used the variants provided by \cite{yu2022point} in our experiments.}

\subsection{Implementation Details}

\paragraph{Pre-training.} For the 3D input, we uniformly sample $N_p=$ 1024, 2048, or 8192 points for accommodating the requirements of different backbones. The inputs of image and text modalities are generated as described in Section \ref{sec:create triplet}. During pre-training, we utilize an advanced version of CLIP, namely SLIP~\cite{mu2022slip}, that shows superior performance as our image-text encoders. 
As mentioned in Section \ref{sec:align representation}, we freeze the image and text encoders and only update the 3D encoder's parameters during pre-training. 
ULIP is trained for 250 epochs.
We use $64$ as the batch size, $10^{-3}$ as the learning rate, and AdamW as the optimizer. %[TODO: add specific settings]

\vspace{-10pt}
\paragraph{Standard 3D Classification.} On ModelNet40, we use the learning rate as 0.00015 and fine-tune our model for 200 epochs, with the batch size as 24 for PointNet++. 
For PointMLP, we set the learning rate as 0.1 and fine-tune the model for 300 epochs, with the batch size as 32. 
%, min lr of 0.005, 300 epochs, batch size of 64. For pointbert, we use a learning rate of 0.0004, for 300 epochs, and a batch size of 32.

On ScanObjectNN, we use the learning rate of 0.03 and finetune for 350 epochs with batch size 32 for PointMLP. 
For PointBERT, we use the learning rate of 0.0002 and finetune for 300 epochs with batch size 32.

\vspace{-10pt}
\paragraph{Zero-Shot 3D Classification.} Following \cite{zhang2022pointclip}, zero-shot 3D classification is conducted by measuring distances between the 3D features of an object and the text features of category candidates. The category that introduces the smallest distance is selected as the predicted category, as shown in Figure \ref{fig:method}. We use our pre-trained models as they are when performing zero-shot classification. There is no finetuning stage involved. We keep using the same prompt strategy as it is during pre-training when constructing text features for each category candidate in this task.

All of our experiments are conducted using PyTorch. Pre-training and finetuning experiments use 8 and 1 A100 GPUs, respectively.

\subsection{Standard 3D Classification}
We demonstrate the effectiveness of ULIP by improving different 3D classification baselines. We follow the original settings of the baselines in our experiments. When applying ULIP, the only difference is that we pre-train the 3D networks under our framework before finetuning them with the labeled point cloud. Since the structure of the 3D backbone is unchanged, our framework does not introduce extra latency during inference time.
For all experiments, we follow the community practice of using OA (Overall Accuracy) and mAcc (Class Average Accuracy) as our evaluation metrics.
%We want to demonstrate the effectiveness of the representation learning using our ULIP framework, so we conduct the experiments to finetune the pre-trained 3d encoder, note that we don't change the model architecture for existing 3d encoders, just take the existing 3d encoders to pre-train in our ULIP framework, and then finetune the pre-trained weights, everything is kept the same except the change of the hyper parameters like learning rate.

% We want to demonstrate the effectiveness of the representation learning using our ULIP framework, so we conduct the experiments to finetune the pre-trained 3d encoder, note that we don't change the model architecture for existing 3d encoders, just take the existing 3d encoders to pre-train in our ULIP framework, and then finetune the pre-trained weights, everything is kept the same except the change of the hyper parameters like learning rate. 

%\subsubsection{Evaluation Metric}
% what metric is used
%We follow the community practice to report OA (Overall Accuracy) and mAcc (class average accuracy) as our evaluation metrics.

\begin{table}[htb]
    \small
    \centering
    \begin{tabular}{lcc}
    \toprule
         Model  &  Overall Acc & Class-mean Acc\\
         \midrule
         PointNet \cite{qi2017pointnet}  & 89.2 & 86.0 \\
         %\hline
         PointCNN \cite{li2018pointcnn}  & 92.2 & -\\
         %\hline
         SpiderCNN \cite{xu2018spidercnn} & 92.4 & -\\
         %\hline
         PointConv \cite{wu2019pointconv} & 92.5 & -\\
         %\hline
         Point Transformer \cite{zhao2021point} & 92.8 & -\\
         %\hline
         KPConv \cite{thomas2019kpconv} & 92.9 & -\\
         %\hline
         DGCNN \cite{wang2019dynamic}  & 92.9 & 90.2 \\
         %\hline
         PCT \cite{guo2021pct}  & 93.2 & -\\
         %\hline
         RS-CNN* \cite{liu2019relation}  & 93.6 & -\\
         %\hline
         GDANet \cite{xu2021learning}  & 93.8 & -\\
         %\hline
         GBNet \cite{qiu2021geometric}  & 93.8 & 91.0 \\
         %\hline
         MTVN \cite{hamdi2021mvtn}  & 93.8 & 92.0 \\
         %\hline
         RPNet \cite{ran2021learning} & 94.1 & -\\
         %\hline
         CurveNet \cite{xiang2021walk} & 94.2 &- \\
         %\hline
         \midrule
         PointNet++(ssg) & 90.7 & -\\
        %  \hline
         PointNet++(ssg) + ULIP & \textbf{93.4} \textcolor{DarkGreen}{\small ($\uparrow 2.7$)} & 91.2 \\
        %  \hline
        %  ULIP + pointnet++ msg & 93.5 & 91.3 \\
        %  \hline
        %  ULIP + pointnet++ 2x width ssg  & 93.8 & 91.8 \\
         %\hline
        %  ULIP + PointNet++ ssg  & \textbf{94.0} & 92.0 \\
         %\hline %8k
         \midrule
         PointBERT  & 93.2 &  -\\
         %\hline %8k w/ voting
        %  point-bert (reproduced) & 93.00 & 93.35 \\
        %  \hline
        PointBERT + ULIP & \textbf{94.1} \textcolor{DarkGreen}{\small ($\uparrow 0.9$)} & - \\
         %\hline %8k
        %  Point-BERT w/ voting & 93.8 &  \\
        %  \hline
        %  ULIP + Point-BERT w/ voting &  94.1 & \\
        %  \hline
         \midrule
         PointMLP  & 94.1 & 91.3 \\
         %\hline
        %  pointMLP w/o voting & 92.99 & 88.81 \\
        %  \hline
         PointMLP + ULIP  &  \textbf{94.3} \textcolor{DarkGreen}{\small ($\uparrow 0.2$)} & \textbf{92.3} \textcolor{DarkGreen}{\small ($\uparrow 1.0$)} \\
         %\hline
         \midrule
         PointMLP*  &  94.5 & 91.4 \\
         %\hline %w/ voting
        %  pointMLP & reproduced & 2k &  & 93.27(wo/ voting) & 90.73 \\
        %  \hline
        PointMLP* + ULIP  &  \textbf{94.7} \textcolor{DarkGreen}{\small ($\uparrow 0.2$)} & \textbf{92.4} \textcolor{DarkGreen}{\small ($\uparrow 1.0$)}\\
         %\hline %w/ voting
         \bottomrule
        %  ULIP & tuned lr and seed & 2k &  & 94.1 (wo/ voting) & 92.23 \\
        %  \hline
    \end{tabular}
    \caption{Standard 3D classification results on ModelNet40. ULIP significantly improves our baselines. Our best number achieves new SOTA. * means a voting technique is applied to the method to boost performance.}
    \label{tab:fintune-modelnet}
    \vspace{-4mm}
\end{table}

\begin{table*}[htbp]
    \small
    \vspace{-4mm}
    \centering
    \begin{tabular}{lccccccc}
         \toprule
         \multirow{2}*{Model} & \multirow{2}*{Modalitiy aligned} & \multicolumn{2}{c}{ALL} & \multicolumn{2}{c}{Medium} & \multicolumn{2}{c}{Hard}
         \\
         \cmidrule(lr){3-4}\cmidrule(lr){5-6}\cmidrule(lr){7-8}
         ~ & ~ & top1 & top5 & top1 & top5 & top1 & top5
         \\
         \midrule
         PointNet++ ssg + ULIP & P+T & 44.9 & 70.3 & 17.2 & 55.0 & 20.3 & 50.1\\
         %\hline
         PointNet++ ssg + ULIP & P+I & 35.3 & 67.2 & 33.6 & 62.4 & 30.1 & 55.1\\
         %\hline
         PointNet++ ssg + ULIP & P+I+T & \textbf{55.7} & \textbf{75.7} & \textbf{35.6} & \textbf{64.4} & \textbf{33.7} & \textbf{55.8}\\
         %\hline
         \midrule
         PointNet++ msg + ULIP & P+T & 48.0 & 63.8 & 17.8 & 42.3 & 21.3 & 40.7\\
         %\hline
         PointNet++ msg + ULIP & P+I & 36.4 & 64.4 & 34.7 & 59.0 & 31.0 & 52.0\\
         %\hline
         PointNet++ msg + ULIP & P+I+T & \textbf{58.4} & \textbf{78.2} & \textbf{36.9} & \textbf{67.2} & \textbf{33.9} & \textbf{59.6}\\
         \midrule
         %\hline
         PointMLP + ULIP & P+T & 52.2 & 73.0 & 23.3 & 60.8 & 18.1 & 52.2\\
         %\hline
         PointMLP + ULIP & P+I & 34.6 & 64.3 & 31.3 & 61.7 & 27.0 & 53.7\\
         %\hline
         PointMLP + ULIP & P+I+T & \textbf{61.5} & \textbf{80.7} & \textbf{43.2} & \textbf{72.0} & \textbf{36.3} & \textbf{65.0}\\
         \midrule
         %\hline
         PointBERT + ULIP & P+T & 44.7 & 66.0 & 19.4 & 49.3 & 14.7 & 39.3\\
         %\hline
         PointBERT + ULIP & P+I & 35.5 & 66.9 & 35.0 & 64.4 & 34.1 & 59.1\\
         %\hline
         PointBERT + ULIP & P+I+T & \textbf{60.4} & \textbf{84.0} & \textbf{40.4} & \textbf{72.1} & \textbf{37.1} & \textbf{66.3}\\
         \bottomrule
    \end{tabular}
    \caption{Analysis of aligning three vs. two modalities on zero-shot 3D classification on ModelNet40. Results show that aligning representations of three modalities always produces better results than two modalities.}
    \label{tab:ablation-modelnet}
    %\vspace{-4mm}
\end{table*}

\begin{table*}[htb]
    \small
    % \vspace{-4mm}
    \centering
    \begin{tabular}{lcccccc}
         \toprule
         \multirow{2}*{Model}  & \multicolumn{2}{c}{ ALL} & \multicolumn{2}{c}{Medium} & \multicolumn{2}{c}{Hard}
         \\
         \cmidrule(lr){2-3}\cmidrule(lr){4-5}\cmidrule(lr){6-7}
         ~  & top-1 & top5 & top-1 & top-5 & top-1 & top-5
         \\
         \midrule
         %\hline
         PointCLIP  & 20.2 & -- & 10.4 & --& 8.3 &-- \\
         %\hline
         \midrule
         PointNet++(ssg) + ULIP& 55.7 & 75.7 & 35.6 & 64.4 & 33.7 & 55.8\\
         %\hline
         PointNet++(msg) + ULIP & 58.4 & 78.2 & 36.9 & 67.2 & 33.9 & 59.6\\
         %\hline
         PointMLP + ULIP & 61.5 & 80.7 & 43.2 & 72.0 & 36.3 & 65.0\\
         %\hline
         PointBERT + ULIP& \textbf{60.4} \textcolor{DarkGreen}{\small ($\uparrow 40.2$)} & 84.0 & \textbf{40.4} \textcolor{DarkGreen}{\small ($\uparrow 30.0$)} & 72.1 & \textbf{37.1} \textcolor{DarkGreen}{\small ($\uparrow 28.8$)} & 66.3\\
         \bottomrule
    \end{tabular}
    \caption{Zero-shot 3D classification on ModelNet40. ULIP-based methods outperform the previous SOTA (PointCLIP) by a very large margin in different evaluation sets.}
    \label{tab:zero-shot-modelnet}
    %\vspace{-4mm}
\end{table*}

\begin{table}[htb]
    \small
    \centering
    \begin{tabular}{lcc}
         \toprule
         \multirow{2}*{Model}  & \multicolumn{2}{c}{ALL}
         \\
         \cmidrule{2-3}
         ~  & top-1 & top-5 
         \\
         \midrule
         PointCLIP & 15.4&-- \\
         %\hline
         \midrule
         PointMLP + ULIP &  44.6 & 82.3\\
         PointNet++(ssg)  + ULIP& 45.6 & 73.8\\
         PointBERT + ULIP  & 48.5 & 79.9\\
         %\hline
         PointNet++(msg) + ULIP & \textbf{49.9} \textcolor{DarkGreen}{\small ($\uparrow 34.5$)} & 78.8\\
         %\hline
         
         %\hline
         
         \bottomrule
    \end{tabular}
    \caption{Zero-shot 3D classification on ScanObjectNN. ULIP-based methods outperform the previous SOTA (PointCLIP) by a very large margin (at least 29.2\% on top-1 accuracy).}% scanobjectnn used 2k points.}
    \label{tab:zero-shot-scan}
    \vspace{-4mm}
\end{table}
%After pre-training, we take the pre-trained 3D backbone networks and further fine-tune them on the two downstream data sets introduced in Section 4.1. 

%For the finetune experiments, we keep the architecture of the 3D encoder unchanged, and initialize the weights using the pre-trained weights. In practice, we find that normally after initialize with the pre-trained weights, the learning rate might not be optimal, and you need to adjust accordingly, if the finetune dataset has a similar distribution to our pre-trained dataset ShapeNet55, normally we find lower the learning rate will help, if the finetune dataset has a domain gap to our pre-trained dataset ShapeNet55, normally we find that increasing the learning rate will help.

\vspace{-10pt}
\paragraph{Experimental Results.}
We present the standard 3D classification performances of our baselines and our methods on ScanObjectNN in Table~\ref{tab:fintune-scan}. As shown, the performances of our baselines are significantly improved by ULIP. Specifically, our framework improves PointBERT and PointMLP significantly by around 3\%.
When we apply ULIP on the strongest backbone, PointMLP, ULIP+PointMLP$\dag$ achieves the new SOTA performance, and outperforms previous SOTA, RepSurf-U(2$\times$), by 3.4\% Overall Accuracy. 

In Table~\ref{tab:fintune-modelnet}, we show the standard 3D classification results on ModelNet40.
Unlike ScanObjectNN, which contains scans of real objects, ModelNet40 is a synthetic dataset thus it is easier for classification. The Overall Accuracy of recent methods is already saturated around $94\%$ on this dataset.
Even in such a scenario, from Table~\ref{tab:fintune-modelnet} we can see that ULIP is still able to improve the Overall Accuracy of all of our baselines. Among them, ULIP+PointMLP* achieves a new SOTA.
For the class-mean accuracy metric, we also observe decent performance improvement when using ULIP and achieves a new SOTA as well.

\subsection{Zero-Shot 3D Classification}
By aligning the 3D representation with text and image representations, ULIP also enables the 3D backbone networks to conduct tasks involving multiple modalities. We evaluate zero-shot 3D classification in this section.

PointCLIP is the first work and the current SOTA for zero-shot 3D classification. It conducts zero-shot 3D classification by first converting a 3D point cloud into 6 orthogonal depth maps, then using CLIP's image encoder to get ensembled depth map features, and finally using CLIP to match text and depth map features for zero-shot classification. We use it as our major baseline and follow its evaluation protocol in this task. For all experiments, we report top-1 and top-5 OA (Overall Accuracy).

\begin{table}[htb]
    \small
    \centering
    \begin{tabular}{lccc}
         \toprule
         \multirow{2}*{Model} & \multirow{2}*{Modality} & \multicolumn{2}{c}{ALL}
         \\
         \cmidrule{3-4}
         ~ & ~ & top-1 & top-5 
         \\
         \midrule%\hline
         PointNet++ ssg + ULIP & P+T & 33.4 & 73.0 \\
         %\hline
         PointNet++ ssg + ULIP & P+I & 35.3 & 72.8\\
         %\hline
         PointNet++ ssg + ULIP & P+I+T & \textbf{45.6} & \textbf{73.8}\\
         %\hline
         \midrule
         PointNet++ msg + ULIP & P+T & 39.2 & 70.4\\
         %\hline
         PointNet++ msg + ULIP & P+I & 34.9 & 71.3 \\
         %\hline
         PointNet++ msg + ULIP & P+I+T & \textbf{49.9} & \textbf{78.8}\\
         %\hline
         \midrule
         PointMLP + ULIP & P+T & 41.3 & 76.1 \\
         %\hline
         PointMLP + ULIP & P+I & 33.6 & 74.7 \\
         %\hline
         PointMLP + ULIP & P+I+T & \textbf{44.6} & \textbf{82.3}\\
         %\hline
         \midrule
         PointBERT + ULIP & P+T & 31.0 & 69.0\\
         %\hline
         PointBERT + ULIP & P+I & 36.3 & 71.3\\
         %\hline
         PointBERT + ULIP & P+I+T & \textbf{48.5} & \textbf{79.9}\\
         \bottomrule
    \end{tabular}
    \caption{Analysis of aligning three vs. two modalities on zero-shot 3D classification on ScanObjectNN. Results show that aligning representations of three modalities always produces better results than two modalities.}
    %, evaluate on all classes, scanobjectnn used 2k points.}
    \label{tab:ablation-scan}
    \vspace{-4mm}
\end{table}

\vspace{-10pt}
\paragraph{Evaluation Sets.}
% describe experimental settings. Input, output and other details
 %The output will be the point cloud to text logits.
 
To perform a fair comparison with PointCLIP, we evaluate zero-shot 3D classification on the entire test sets of both ModelNet40 and ScanObjectNN. We refer to this set as \emph{ALL}.

Besides, we notice that there are some common classes between our pre-train dataset, ShapeNet55, and ModelNet40. Evaluations on these common classes might introduce an unfair comparison of zero-shot performance.
To deal with this issue, we create two more sets in ModelNet40, referred to as \emph{Medium} and \emph{Hard} for evaluation.

\noindent\textit{Medium set}: We remove the ModelNet40 categories whose exact category names exist in our pre-training category list.

\noindent\textit{Hard set}: In the ``Medium'' category list, there are still some category names that are synonyms to the pre-training categories, 
such as 'cup' vs. 'mug' and 'chair' vs. 'stool.' Therefore, for the ``Hard'' ModelNet40 category list, we remove the categories from the ``Medium'' list with semantically similar counterparts in pre-training categories. 
%(TODO: how did you decide whether two class names are semantically similar here?)
% what metric is used

\begin{figure}[htb]
    \centering
    \includegraphics[width=1.0\linewidth]{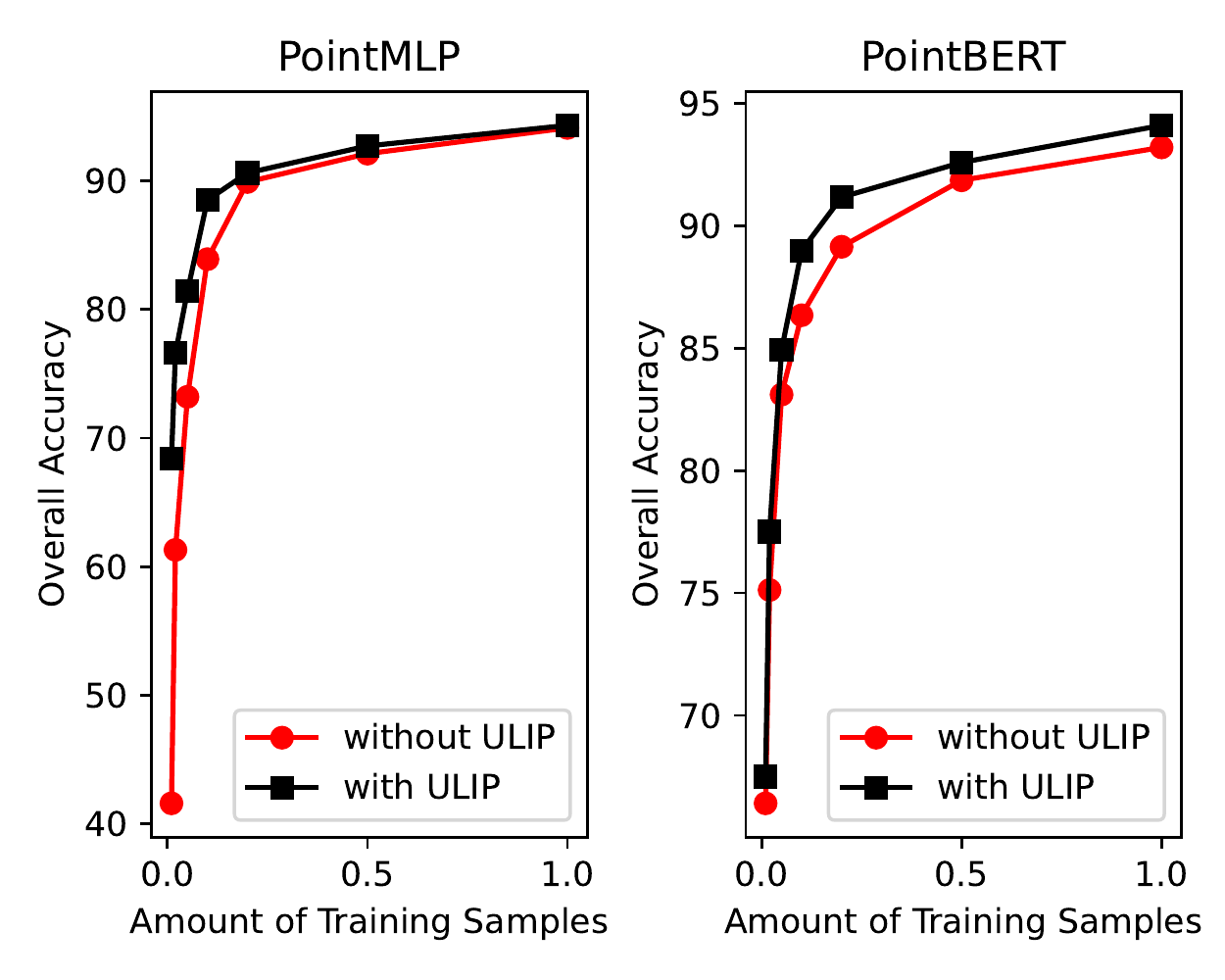}
    %\vspace{-8mm}
    \caption{Data efficiency comparison. The X axis indicates the percentage of samples used for training and Y axis denotes the overall accuracy. Both PointMLP and PointBERT are significantly improved when pre-training with ULIP. }
    \label{fig:data-efficiency}
    %\vspace{-6mm}
\end{figure}
\begin{figure*}[htb]
    \centering
    \includegraphics[width=1.0\linewidth]{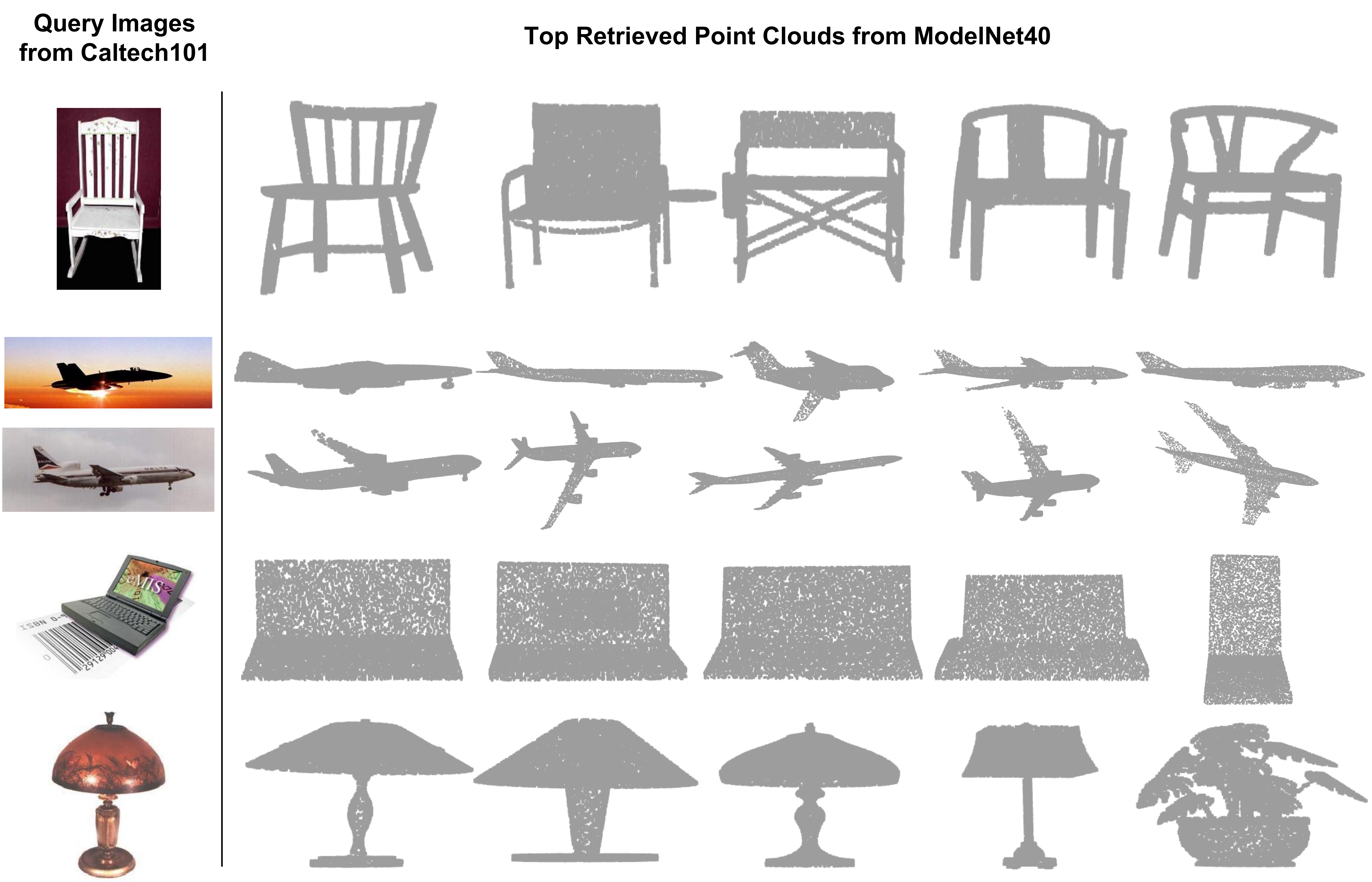}
    %\vspace{-20pt}
    \caption{Qualitative results of real image to point cloud retrieval. Query images are from Caltech101, and point clouds are from ModelNet40. We show the top-5 retrieved point cloud models, ranked in order. The results demonstrate the retrieval capability of our model.}
    \label{fig:image_pc_retrieval}
    %\vspace{-6mm}
\end{figure*}
\paragraph{Experimental Results.}
% result goes here
We present the zero-shot 3D classification results on ModelNet40 in Table~\ref{tab:zero-shot-modelnet} and the results on ScanObjectNN in Table~\ref{tab:zero-shot-scan}. We can see that all ULIP-based methods significantly outperform our major baseline, PointCLIP, by a large margin in every evaluation set.
Specifically, on the \emph{Hard set}, our best performing method, ULIP + PointBERT, outperforms PointCLIP by around 29\% in top-1 accuracy. It also indicates that the superior performance of ULIP-based methods is not caused by pre-training the model on exact/similar categories as the target categories.
Instead, it suggests that aligning the representations of different modalities can benefit the recognition of rare categories in general. Results in Table \ref{tab:zero-shot-scan} demonstrate that ULIP-based methods consistently surpass PointCLIP on real scanned objects. Furthermore, all of the 3D backbones outperform the SOTA zero-shot method, PointCLIP, by $\sim$30\% with the help of our ULIP framework.

\subsection{Analyses}
\label{sec:more analyses}

%\vspace{-5pt}
\paragraph{Align Representations, Three Modalities or Two?}
As described in Eq. \ref{eq:final loss}, ULIP by default aligns the 3D representation with both the text and image representations during pre-training.
We wonder to what extent ULIP will still work if we align the 3D representation to only the text feature or image feature alone.
In this section,
we conduct an ablation study for ULIP by aligning two rather than three modalities in zero-shot settings.
Results are shown in Table~\ref{tab:ablation-scan} and Table~\ref{tab:ablation-modelnet} for ScanObjectNN and ModelNet40 datasets, respectively. As we can see in both tables, aligning the 3D modality with both text and image modalities consistently achieves the best performance compared to aligning with either image or text modality in every scenario with each baseline.

\vspace{-10pt}
\paragraph{Data Efficiency.}
Model pre-training could potentially reduce the demand for labeled data during fine-tuning in downstream tasks. We validate the data efficiency of ULIP by comparing it with baselines under a varying number of fine-tuning samples. The comparison results are shown in Figure \ref{fig:data-efficiency}. As shown in Figure \ref{fig:data-efficiency} (left), PointMLP's performance is largely improved in the low data regime when pre-trained under the ULIP framework. When we compare PointBERT with PointMLP baselines (two red lines in the two sub-figures), we observe that PointBERT performs better than PointMLP when using less than 20\% training data. This is because of that the PointBERT model itself is pre-trained on ShapeNet55. Although both ULIP and PointBERT are pretrained on ShapeNet55, ULIP still improves PointBERT by a clear margin, as shown in Figure \ref{fig:data-efficiency} (right).

%-------------------------------------------------------------------------

\subsection{Cross-Modal Retrieval}
As mentioned in Section~\ref{sec:intro}, one of the benefits of ULIP is that it enables more cross-modal downstream tasks. Here, we qualitatively show the potential of using ULIP to conduct real image to point cloud retrieval. 

We use our pre-trained ULIP with PointBERT as the 3D encoder directly. We conduct a small-scale experiment with real images from Caltech101 \cite{fei2004learning} and use the images to retrieve 3D point clouds from around 2.5k samples over 40 categories in ModelNet40. In Figure \ref{fig:image_pc_retrieval}, we show the top-5 retrieved 3D point cloud models (ranked in order) using image examples from categories of \emph{chair}, \emph{airplane}, \emph{laptop} and \emph{lamp}. The results show encouraging signs that our pre-trained model has learned meaningful features across image and 3D point cloud modalities. Surprisingly, the top-1 retrieved 3D models have the closest appearance to the query images compared to other retrieved 3D models. For example, when we use images from different aircraft types (fight and airliner) for retrieval (2nd and 3rd rows), the retrieved top-1 point clouds maintain the subtle difference of the query images.

\section{Conclusions}
\label{sec:conclusion}
We propose ULIP, a pre-training framework that aligns multiple modalities of images, language, and point clouds in the same feature space. We take advantage of the pre-trained text and image encoders and improve different 3D encoders using our framework. Experiments results show that ULIP can effectively improve representations of 3D backbones. Our method achieves state-of-the-art performance in both zero-shot and standard 3D classification tasks, and our qualitative results show that ULIP has promising potential for cross-modal retrieval applications.

% \begin{figure}[tb]
%     \centering
%     \includegraphics[width=1.0\linewidth]{cvpr/subsample_pointmlp.png}
%     %\vspace{-8mm}
%     \caption{Data efficiency comparison between PointMLP and ULIP+PointMLP. X axis indicates the percentage of data used for fine-tuning. Results are on ModelNet40.}
%     \label{fig:data efficiency pointmlp}
%     %\vspace{-6mm}
% \end{figure}

% \begin{figure}[tb]
%     \centering
%     \includegraphics[width=1.0\linewidth]{cvpr/pointbert_subsample.png}
%     %\vspace{-8mm}
%     \caption{Data efficiency comparison between PointBERT and ULIP+PointBERT. X axis indicates the percentage of data used for fine-tuning. Results are on ModelNet40.}
%     \label{fig:data efficiency pointbert}
%     %\vspace{-6mm}
% \end{figure}

% \begin{figure}[tb]
%     \centering
%     \includegraphics[width=1.0\linewidth]{cvpr/pointbert_data_efficiency.pdf}
%     %\vspace{-8mm}
%     \caption{}
%     \label{fig:data efficiency}
%     %\vspace{-6mm}
% \end{figure}

% \begin{figure}[tb]
%     \centering
%     \includegraphics[width=1.0\linewidth]{cvpr/pointmlp_data_efficiency.pdf}
%     %\vspace{-8mm}
%     \caption{}
%     \label{fig:data efficiency}
%     %\vspace{-6mm}
% \end{figure}

%-------------------------------------------------------------------------

%%%%%%%%% REFERENCES
{\small
\bibliographystyle{ieee_fullname}
\bibliography{main}
}

\clearpage

\appendix

\section{Appendix}
\label{sec:appendix}

\subsection{PointNeXt Backbone Experiments}
\noindent\textbf{PointNeXt}\cite{qian2022pointnext} is a concurrent work which proposes a lightweight backbone based on PointNet++ and in particularly it gives promising results on the ScanObjectNN benchmark. In order to demonstrate the effectiveness of our ULIP on this most recent backbone, we pre-train PointNeXt using ULIP, and use the pre-trained weights to finetune on the ScanObjectNN dataset.

As shown in Table \ref{tab:fintune-scan}, ULIP significantly improves PointNeXt in both Overall Accuracy and Class-mean Accuracy.

\begin{table}[htb]
    %\small
    \centering
    \begin{tabular}{lcc}
    \toprule
         Model& Overall Acc & Class-mean Acc \\
         \midrule
        %  pointMLP & reproduced & 1k & 85.22 & 84.13 \\
        %  \hline
         PointNeXt* \cite{qian2022pointnext} &  87.4 & 85.8 \\
         %\hline
         PointNeXt + ULIP &  \textbf{89.2} \textcolor{DarkGreen}{\small ($\uparrow 1.8$)} & \textbf{88.0} \textcolor{DarkGreen}{\small ($\uparrow 2.2$)} \\
         %\hline
         \midrule
         PointNeXt \dag * & 87.5  & 85.9 \\
         %\hline
         PointNeXt \dag + ULIP &  \textbf{89.7} \textcolor{DarkGreen}{\small ($\uparrow 2.2$)} & \textbf{88.6} \textcolor{DarkGreen}{\small ($\uparrow 2.7$)} \\
         %\hline
         \bottomrule
    \end{tabular}
    \caption{3D classification results on ScanObjectNN for PointNeXt. \dag indicates a model uses 2K sampled points and all others use 1K sampled points. * indicates it's reproduced result.}
    \label{tab:fintune-scan}
    %\vspace{-4mm}
\end{table}

\subsection{Details of Evaluation Sets in  Zero Shot Classification}
\noindent When evaluating zeroshot classification, we notice that there are some common classes between our pre-train dataset, ShapeNet55, and ModelNet40. Evaluations on these common classes might introduce an unfair comparison of zeroshot performance. Therefore, we introduced three different validation sets for evaluating our models and our baselines on ModelNet40.

\noindent\textbf{All Set}: Includes all the categories in ModelNet40 as shown in Table \ref{tab:ModelNet40-All-Set}.

\begin{table}[htb]
    \small
    \begin{tabular}{ccccc}
        \toprule
        airplane & bathtub & bed & bench & bookshelf \\
        \midrule
        bottle & bowl & car & chair & cone \\
        \midrule
        cup& curtain& desk& door& dresser \\
        \midrule
        flower\_pot& glass\_box& guitar& keyboard& lamp \\
        \midrule
        laptop& mantel& monitor& night\_stand& person \\
        \midrule
        piano& plant& radio& range\_hood& sink \\
        \midrule
        sofa& stairs& stool& table& tent \\
        \midrule
        toilet& tv\_stand& vase& wardrobe& xbox \\
        \bottomrule
    \end{tabular}
    \caption{ModelNet40 All Set.}
    \label{tab:ModelNet40-All-Set}
\end{table}

\noindent\textbf{Medium Set}: We remove categories whose
exact category names exist in our pre-training dataset. The resulting categories in this set is shown in Table \ref{tab:ModelNet40-Medium-Set}.

\begin{table}[htb]
    \small
    \begin{tabular}{ccccc}
        \toprule
        cone& cup& curtain& door& dresser \\
        \midrule
        glass\_box& mantel& monitor& night\_stand& person \\
        \midrule
        plant& radio& range\_hood& sink& stairs \\
        \midrule
        stool& tent& toilet& tv\_stand& vase \\
        \midrule
        wardrobe& xbox \\
        \bottomrule
    \end{tabular}
    \caption{ModelNet40 Medium Set.}
    \label{tab:ModelNet40-Medium-Set}
\end{table}

\noindent\textbf{Hard Set}: We remove both extract category names and their synonyms in our pre-training dataset. The final \emph{Hard Set} is shown in Table \ref{tab:ModelNet40-Hard-Set}

\begin{table}[htb]
    \small
    \begin{tabular}{ccccc}
        \toprule
        cone& curtain& door& dresser& glass\_box \\
        \midrule
        mantel& night\_stand& person& plant& radio \\
        \midrule
        range\_hood& sink& stairs& tent& toilet \\
        \midrule
        tv\_stand& xbox \\
        \bottomrule
    \end{tabular}
    \caption{ModelNet40 Hard Set.}
    \label{tab:ModelNet40-Hard-Set}
\end{table}

\subsection{Indoor 3D Detection Experiments}
In order to show our potential on 3D scene applications, we conduct experiments on ScanNet-v2 dataset and benchmark 3D detection performance based on one of SOTA 3D detection frameworks, Group-Free-3D \cite{liu2021group}. In our setting, we use the Group-Free-3D basic model and observe significant improvements as shown in Table \ref{tab:GroupFree3D experiments}.

\begin{table}[htb]
    \small
    % \vspace{-4mm}
    \centering
    \begin{tabular}{lcc}
         \toprule
         % \multirow{2}*{Model}  & \multicolumn{2}{c}{ALL}
         % \\
         % \cmidrule{2-3}
         ~ & mAP@0.5 & mAP@0.5 Averaged
         \\
         \midrule
         Group-Free-3D & 48.9 & 48.4\\
         %\hline
         \midrule
         Group-Free-3D + ULIP & 50.2 \textcolor{DarkGreen}{\small ($\uparrow 1.3$)} & 49.6 \textcolor{DarkGreen}{\small ($\uparrow 1.2$)}\\
         \bottomrule
    \end{tabular}
    % \vspace{-4mm}
    \caption{Experiments on indoor 3D Detection. We use Group-Free-3D basic model as our detection framework, and we follow the same metric computation as in \cite{liu2021group}.}
    \label{tab:GroupFree3D experiments}
    %\vspace{-4mm}
\end{table}

\end{document}